%% file: principled_grasp_selection.tex
\documentclass[conference,table]{IEEEtran}
\IEEEoverridecommandlockouts
\usepackage{graphicx}
\usepackage{tikz}
\usetikzlibrary{positioning,shapes,shadows,arrows}
\usetikzlibrary{matrix}
\usetikzlibrary{trees}
\usepackage[labelformat=simple]{subfig}
\usepackage[algoruled,vlined,linesnumbered]{algorithm2e}
\usepackage{amssymb,amsmath}
\usepackage{url}
\usepackage{multirow}
\usepackage{booktabs}
\usepackage{chngcntr}

\input{latex_config}

\begin{document}
\shrinka

\title{Grasping for a Purpose: Using Task Goals for Efficient Manipulation Planning}
\author{Ana Huam\'{a}n Quispe$\quad\quad$ Heni Ben Amor $\quad\quad$ Henrik I. Christensen $\quad\quad$ M. Stilman$\dagger$
\thanks{Institute for Robotics and Intelligent Machines, Georgia Institute
of Technology, Atlanta, GA 30332, USA. {\tt\small ahuaman3@gatech.edu, hbenamor@cc.gatech.edu, hic@cc.gatech.edu} } }
\maketitle

\begin{abstract}
\input{abstract}
\end{abstract}

\section{Introduction}
\label{sec:Introduction}
\input{introduction}

\begin{figure*}[t!]
\begin{center}
\includegraphics[width=0.95\textwidth]{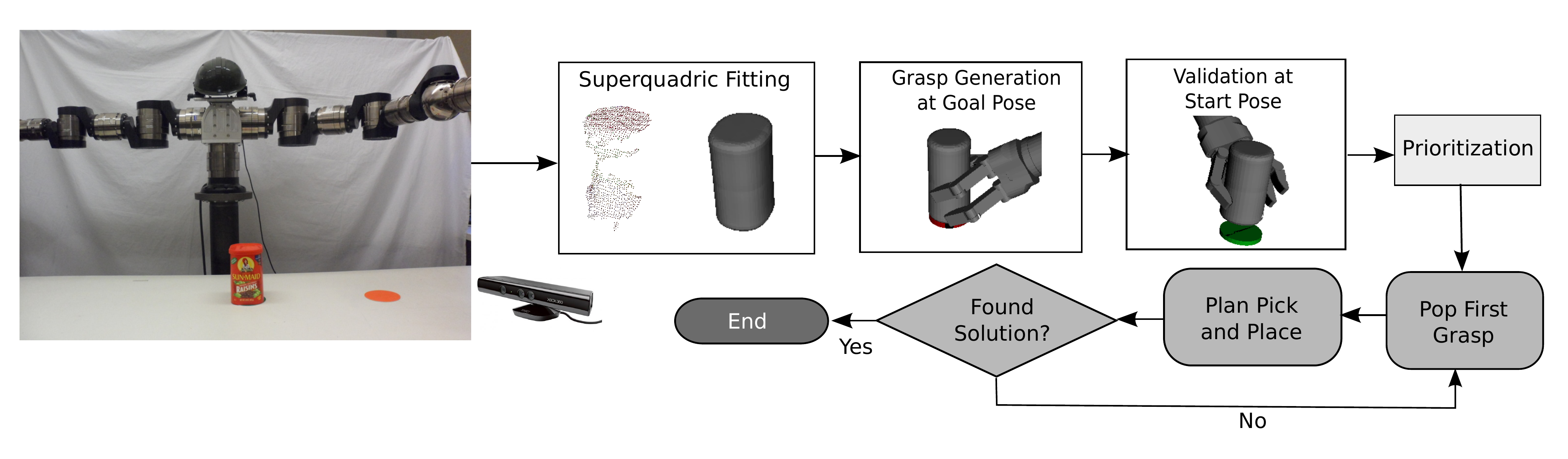}
\caption{Manipulation planning pipeline: a partial point cloud of the object is first analyzed for symmetries and then turned into
a superquadric representation. Grasps at the goal and the start position are generated and then prioritized according to end-comfort. Potential grasps are then analyzed within the plan and then executed.}
\label{fig:unimanualPipeline}
\end{center}
\vspace{-2.0em} 
\end{figure*}

\section{Related Work}
\label{sec:RelatedWork}

\input{relatedWork}

\section{Problem Definition and Assumptions}
\label{sec:ProbDefAssump}

\input{defAssumptions}

\section{Grasp Generation for Unknown Objects}
\label{sec:GraspGeneration}
\input{unimanualTasks}

\section{Grasp Prioritization}
\label{sec:GraspPrioritization}
\input{graspPrioritization}

\section{Experiments and Results}
\label{sec:experiments}
\input{comparativeStudies}



\section{Conclusion}
\label{sec:Conclusion}
\input{conclusion}

\bibliography{principled_grasp_selection}
\bibliographystyle{plain}

\end{document}

%% file: latex_config.tex
\usepackage{cite}

\newcommand{\shrinka}{\def\baselinestretch{1.0}\large\normalsize}

\SetCommentSty{mycommfont}

\newcommand{\Arm}[1]{$\mathcal{A}_{#1}$}
\newcommand{\Hand}[1]{$\mathcal{H}_{#1}$}
\newcommand{\manip}[1]{$m(#1)$}

\newcommand{\SampleSet}{\ensuremath{\mathcal{S}}}

\newcommand{\ep}[1]{$\epsilon_{#1}$}
\newcommand{\param}{$p_{sq}$}

\newcommand{\GraspSet}[1]{$\mathcal{G}_{#1}$}
\newcommand{\grasp}[1]{\ensuremath{\mathbf{g}_{#1}}}

\newcommand{\Tf}[2]{$^{#1}T_{#2}$}

\newcommand{\conf}[1]{\ensuremath{\mathbf{q}_{#1}}}

\newcommand{\Robot}[1]{$\mathcal{R_{#1}}$}
\newcommand{\Object}{$\mathcal{O}$}

\def\aref#1{Algorithm~\ref{#1}}
\def\fref#1{Figure~\ref{#1}}
\def\tref#1{Table~\ref{#1}}

\usetikzlibrary{mindmap,trees}

\definecolor{bgcolor}{rgb}{0.98, 0.98, 1.0}
\definecolor{bgcolorGray}{rgb}{0.8, 0.8, 0.8}
\definecolor{lightGray}{rgb}{0.9,0.9,0.9}
\newenvironment{tikzbox}
{\begin{tikzpicture}
    \node[fill=bgcolor,thick,rounded corners=1em, draw=black,minimum width=0.9\linewidth,minimum height=2em]}
{;\end{tikzpicture}
}

\tikzset{
  basic/.style = {draw,text width=2cm,drop shadow,rectangle},
  root/.style = {basic,rounded corners=2pt,thin,align=center,fill=gray!50},
  level 2/.style = {basic,rounded corners=6pt,thin,align=center,fill=gray!25,
    text width=6em},
  level 3/.style={basic,thin,align=center,fill=pink!50,text width=4.5em,sibling distance=2em}
}

\colorlet{tableheadcolor}{gray!25}
\newcommand{\headcol}{\rowcolor{tableheadcolor}}
\colorlet{tablerowcolor}{gray!10}

\newcommand{\topline}{\arrayrulecolor{black}
  \specialrule{0.1em}{\abovetopsep}{0pt}%
  \arrayrulecolor{tableheadcolor}\specialrule{\belowrulesep}{0pt}{0pt}%
  \arrayrulecolor{black}}
\newcommand{\midline}{\arrayrulecolor{tableheadcolor}
  \specialrule{\aboverulesep}{0pt}{0pt}%
  \arrayrulecolor{black}\specialrule{\lightrulewidth}{0pt}{0pt}%
  \arrayrulecolor{white}\specialrule{\belowrulesep}{0pt}{0pt}%
  \arrayrulecolor{black}}
\newcommand{\bottomline}{\arrayrulecolor{white}\specialrule{\aboverulesep}{0pt}{0pt}%
            \arrayrulecolor{black}\specialrule{\heavyrulewidth}{0pt}{\belowbottomsep}}%

%% file: abstract.tex
In this paper we propose an approach for efficient grasp selection for
manipulation tasks of unknown objects. Even for simple tasks such as pick-and-place, 
a unique solution
is rare to occur. Rather, multiple candidate grasps must be considered and
(potentially) tested till a successful, kinematically feasible path is found.
To make this process efficient, the grasps should be ordered such that 
those more likely to succeed are tested first. We propose to use 
\textit{grasp manipulability}
as a metric to prioritize grasps. We present results of simulation experiments
which demonstrate the usefulness of our metric. Additionally, we present
experiments with our physical robot performing simple manipulation tasks
with a small set of different household objects.

%% file: introduction.tex
The ability to grasp objects in order to accomplish a task  
is one of the hallmarks of human intelligence. Numerous psychological studies show that humans grasp selection depends on the \textit{goal} to be \textit{accomplished} \cite{loucks2012role}. Decision making during
grasping is therefore not only based on stability during manipulation, but also based on task requirements. If a specific grasp does not facilitate the execution of the upcoming sub-tasks, it is omitted from the reasoning process. 

In contrast to that, research on robot grasp synthesis has been tilted towards optimizing stability metrics only. A prominent approach is to generate a set of  physically stable grasps, one of which is then selected by the high-level planner.
If a high-level task planner cannot achieve the goals of the task, it has to back track and try a different grasp. Since no information is flowing between high-level planning and lower-level grasp generation, a large number of grasps may have to evaluated. If the required grasp is not within the optimized set of candidates, the entire task will fail. 

In this paper, we introduce a method for manipulation planning which uses foresight to identify tasks constraints. Constraints extracted from subsequent sub-tasks are used to synthesize grasps that facilitate overall task completion. Our goal is to derive a fast planning algorithm that can efficiently generate manipulation sequences for \emph{previously unseen} objects. These latter properties, hence, allow a robot to perform manipulation tasks in new environments without resorting to prior 3D models of the object or pre-calculated grasp sets. This ability to generalize is realized by using a super-quadric representation of objects. We show how super-quadrics can be extracted from a single depth image and how they can be used to generate a large set grasp candidates.


\begin{figure}[t]
\begin{center}
\includegraphics[width=0.4\textwidth]{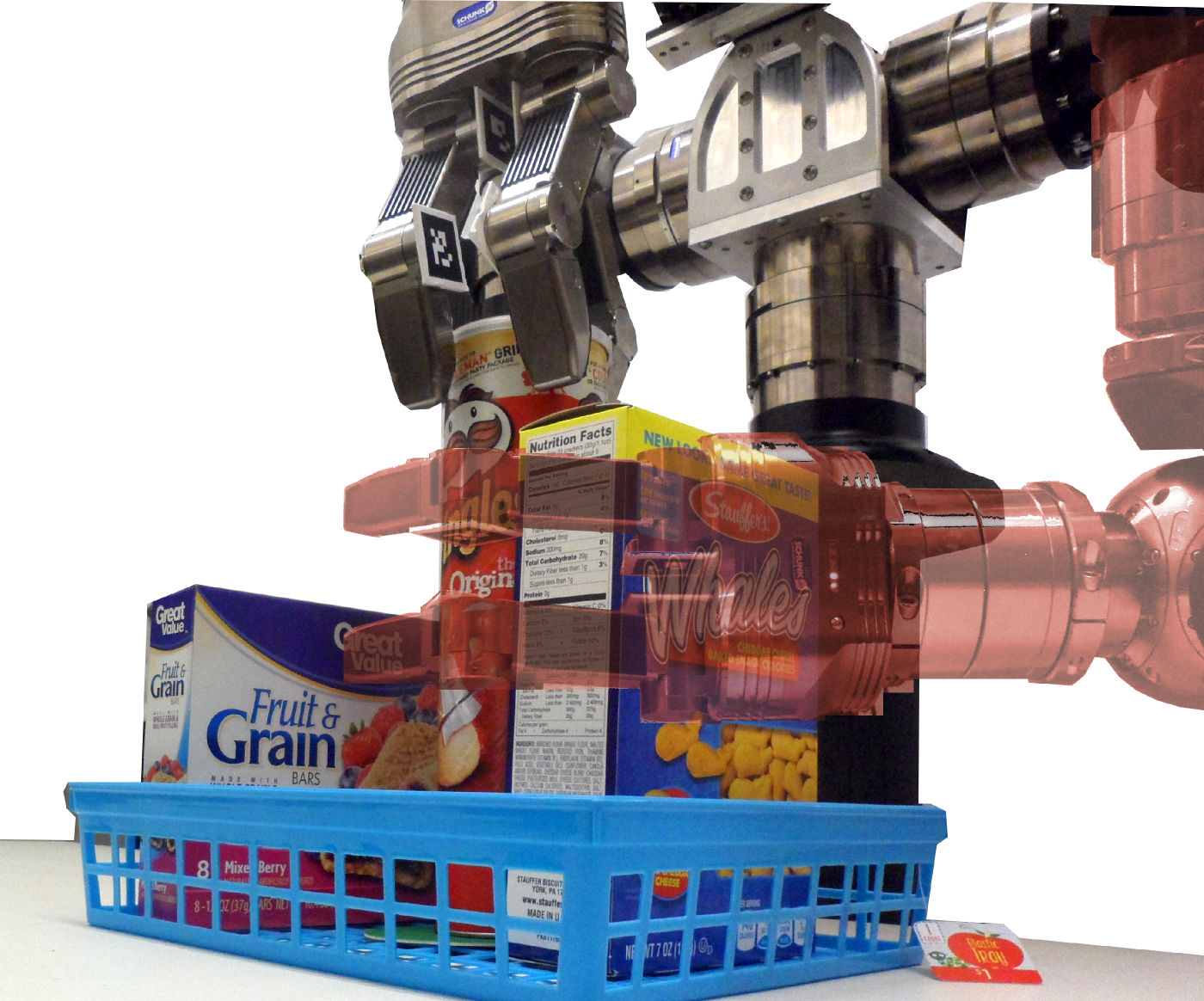}
\caption{Example of grasp selection based on goal constraints: The goal  location of the chips box is surrounded by nearby objects, hence an overhead grasp must be selected for manipulation. If not for the obstacles present, a different grasp (overlayed in red) from the side would be easier to execute.}
\label{fig:coverImage}
\end{center}
\vspace{-2.0em}
\end{figure}


Short planning times are realized by introducing a heuristics that efficiently guides the search by incorporating arm kinematics. Inspired by the \textit{end-comfort effect}\cite{rosenbaum1996cognition} in humans, grasps are preferred which lead to a comfortable arm configuration at the end of a task. We borrow ideas from this work and propose to use a metric based on manipulability as a measurement of end-comfort.

The contributions of this paper are threefold, namely (1) a framework for online grasp planning that incorporates future task constraints into the grasp synthesis process, (2) an efficient grasp generation approach based on super-quadrics that works with previously unseen objects, (3) the end-comfort heuristics for efficient search during manipulation planning.


The rest of this paper is organized as follows: Section \ref{sec:RelatedWork} present relevant
work in the area of grasp synthesis and grasp selection. Section \ref{sec:GraspGeneration} presents
 our grasp generation method using a primitive-based approach and in Section \ref{sec:GraspPrioritization} we introduce our manipulability-based strategies
to prioritize the generated grasps. Section \ref{sec:experiments} shows the results
of the comparisons in simulation and the metrics
we used to compare their performance. Finally, we present the application of our
approach in our physical robot. We conclude this paper with section \ref{sec:Conclusion},
where  we provide some discussion
regarding future work, and the advantages and shortcomings of our approach.

%% file: relatedWork.tex
In this section we review work concerning grasp synthesis 
and grasp selection. For a more detailed review of previous
research in the area, we suggest the interested reader to
consult the excellent reviews from Bohg\cite{bohg2014data} and 
Sahbani\cite{sahbani2012overview}. 

Pioneering work on grasp selection was developed by Cutkosky \cite{cutkosky1989grasp}, who
observed that humans select grasps in order to satisfy 3 main types of constraints:
Hand, object and task-based constraints. As pointed out by Bohg et al. in \cite{bohg2014data},
there is little work on task-dependent grasping when compared to work focused on the first
two constraints. Hence, the main goal for a planner is to find a grasp such that the 
robot can approach the object and execute the said grasp, without further regard of what the
robot will do once the object is picked.


Grasp generation methods vary widely depending on the assumptions considered. 
In the case of grasp planning for known objects, Ciocarlie et al.\cite{ciocarlie2007dimensionality} presented the concept of eigengrasps, which was exploited to generate candidate grasps searching in a 
low dimensional hand posture space using their GraspIt! simulator.
Diankov generated grasps by sampling the surfaces of object meshes and using the normals
at the sample points to guide the approach direction of the hand\cite{diankov2008openrave}. Approaches using primitive 
representations were also proposed such that the grasp generation depends on
the particular primitive characterization: Miller et al.\cite{miller2003automatic} proposed to use a set 
of primitive shapes (cylinder, box, ball) to decompose complex objects. Huebner and Kragic\cite{huebner2008selection} used bounding boxes, Przybylski et al. proposed the Medial Axis representation\cite{przybylski2010unions}, Goldfeder et al.\cite{goldfeder2007grasp} used superquadrics due to their versatility to express
different geometry types with only 5 parameters.

In all the cases mentioned, the grasps are generated
offline and stored in a database for future use. These grasps are usually 
ranked based on their force-closure properties, which theoretically express the 
robustness and stability of a grasp. One of the most popular metrics ($\epsilon$) was proposed
 by Ferrari and Canny \cite{ferrari1992planning}. However, it has been noted by different authors that
 analytical metrics do not guaranteee a stable grasp when executed in a real robot. 
This can be explained by the fact that these classical metrics consider assumptions 
that don't always hold true in real scenarios (dynamics, perceptual and modelling inaccuracies, 
friction conditions). On the other hand, studies that consider human heuristics to guide 
grasp search have shown remarkable results, outperforming classical approaches. In
\cite{balasubramanian2014physical}, Balasubramanian observed that when humans kinestetically teach a robot how to grasp objects, they strongly tend to align the robotic hand along 
one of the object's principal axis, which later results in more robust grasps. The author
termed \textit{skewness} to the metric measuring the axis deviation. In \cite{przybylski2011human}, Przybylski et al. combine the latter metric with $\epsilon$ and use it to rank
grasps produced with GraspIt!. Berenson et al.\cite{berenson2007grasp} proposed a score combining
3 measures: $\epsilon$, object clearance and the robot relative position to the object.

In this work we are interested in manipulation of unknown objects. Multiple approaches
of this kind have flourished during the last few years, 
particularly due to the advent of affordable RGB-Depth sensors. Since the 3D information is
partial and noisy, classical approaches to grasp generation cannot be directly used. Rather,
most of the current work uses heuristics to guide grasp generation based on local representation
of the object geometry features (or global features if the object shape is approximated). In
\cite{hsiao2010contact}, Hsiao et al. use the bounding box of the object segmented pointcloud
to calculate grasp approach directions using a set of heuristics. We should notice that for
most of these approaches, their effectiveness can only be verified empirically.

As we mentioned at the beginning of this section, the metrics we discussed above do not consider
the task to be executed \textit{after} the grasp is achieved. Some authors, however, have
investigated this issue at some level. In pioneering work, \cite{li1988task} Li and Sastry
proposed the concept of the  \textit{task ellipsoid}, which maximizes the forces to be
applied in the direction of the task. More recently, Pandey et al. \cite{pandey2012towards} proposed a
framework to select a grasp such that the object grasped can be manipulated in a human-robot
interaction scenario in which the goal pose of the object is not entirely constrained.

Finally, although one of our main concerns is to select a grasp that is suitable for the
task to be executed, we also consider important to use a grasp that allows for a simple, 
easy arm execution. Interestingly, the problem of grasp planning is usually considered isolated
from arm planning. In some recent work, Vahrenkamp et al. proposed Grasp-RRT \cite{vahrenkamp2012simultaneous} in order to perform
both grasp and arm planning combined. In a similar vein, Roa et al. also proposed an approach
that solve both problems simultaneously \cite{roa2014integrated}. Both approaches focus on \textit{reaching tasks}. Our approach 
tackles pick-and-place tasks in which reaching is only the first half of the solution (\textit{object 
placing} being the second). We make use of our proposed heuristics to solve the complete pick-and-place problem in a manner
as efficient as possible by means of grasp prioritization.

%% file: defAssumptions.tex
Our problem description can be explained as follows:
Given a bimanual manipulator \Robot{} and a simple object \Object,
the manipulation task consists on transporting \Object~from a given 
start pose \Tf{w}{s} to a final pose \Tf{w}{g}. 

\begin{figure}[t!]
\begin{center}
\includegraphics[width=0.4\textwidth]{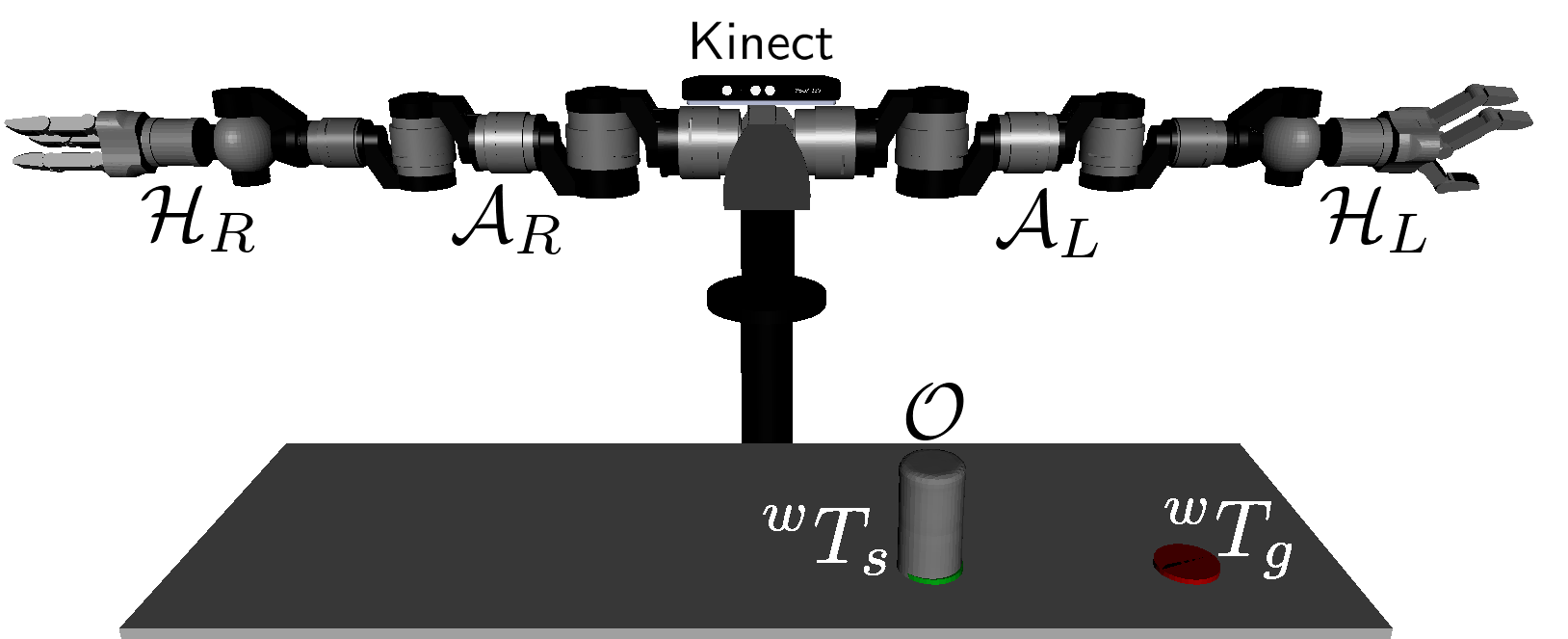} 
\caption{System setup and problem description.}
\label{fig:probDescription}
\end{center}
\end{figure}

\fref{fig:probDescription} depicts the problem described.
The following constraints are considered:

\begin{itemize}
\item{A 3D model of \Object~is not available beforehand.}
\item{A one-view pointcloud of the scenario is available from
the Kinect sensor mounted on top of the robot shoulders.}
\item{Each limb of \Robot{} consists of a 7-DOF arm (\Arm{L},
\Arm{R}) and a 3-fingered hand (\Hand{L}, \Hand{R}). A semi-analytical IK solver
is available for \Arm{L} and \Arm{R} }
\end{itemize}

In the following sections  we will describe our basic approach for problems
in which only the use of one arm is required to solve
the manipulation task described.

%% file: unimanualTasks.tex
As our problem description stated, our approach must
find a plan such that \Object~can be grasped at \Tf{w}{s}, transported
and finally repositioned at \Tf{w}{g}. Our approach consists on
4 main steps, shown in \fref{fig:unimanualPipeline}. The first 3 steps 
(object fitting, grasp generation and grasp validation) will be explained 
in the rest of this section.

\subsection{Object Representation using Superquadrics}
\label{subsec:objRep}
Requiring complete 3D models of objects before grasp synthesis
severely limits the application domains of robot manipulation. Modern
depth cameras partly solve the problem, since they allow the
robot to estimate the surface of an object. Yet, since 
the point clouds are acquired from a specific perspective, 
they only hold partial shape information about the visible frontal part. 
To fill any gaps and produce a complete point cloud, multiple images 
can be acquired by either iteratively moving the camera or the object. 
This process is time-consuming and introduces new challenges such as 
the precise matching of the individual point clouds of each view.

To solve this problem, we use a super-quadric representation of objects
and reason about symmetry in order to infer the shape of any invisible 
part. Superquadrics are a family of geometric shapes that can
represent a wide range of diverse objects using a limited set of parameters.
Superquadrics can be expressed with their implicit equation:

\begin{equation}
\left( \left(\dfrac{x}{a}\right)^{\frac{2}{\epsilon_{2}}} + \left(\dfrac{y}{b}\right)^{\frac{2}{\epsilon_{2}}} \right) ^{\frac{\epsilon_{2}}{\epsilon_1} } + \left(\dfrac{z}{c}\right)^{\frac{2}{\epsilon_{1}}} = 1
\label{eq:impSQeq}
\end{equation}

In our approach, we generate a super-quadric representation using a 
single depth image. Fitting of the parameters can be performed 
online by minimizing the difference between the model and 
the partial point cloud~\cite{huaman2015sq}. However, since only
one side of the object is visible, a standard approach to fitting
will result in erroneous approximations of the object. To
reproduce the entire shape from a partial point cloud, we added
an additional pre-processing step to the superquadric fitting process. 
Instead of using the original point cloud as input, we generate 
a mirrored version by finding an optimal symmetry plane~\cite{huaman2015sq}.
The goals of this step is to exploit symmetries to infer invisible 
parts of an object.

The output of this process for a given object \Object~consists on a transformation \Tf{w}{o}
in world coordinates and the parameters, 
\begin{center}
\param $=$ $\{$ $a$,~$b$,~$c$,~\ep{1},~\ep{2} $\}$
\end{center}

defining its approximated geometry. A good number of household objects can be easily described with generic shapes  such as boxes, cylinders and ovoids, for which we can further bound 
the shape parameters considered: \[ \epsilon_{1} \text{,} \epsilon_{2} \in [0.1,1.9] \]

\begin{figure}[t]
\begin{center}
\includegraphics[width=0.9\linewidth]{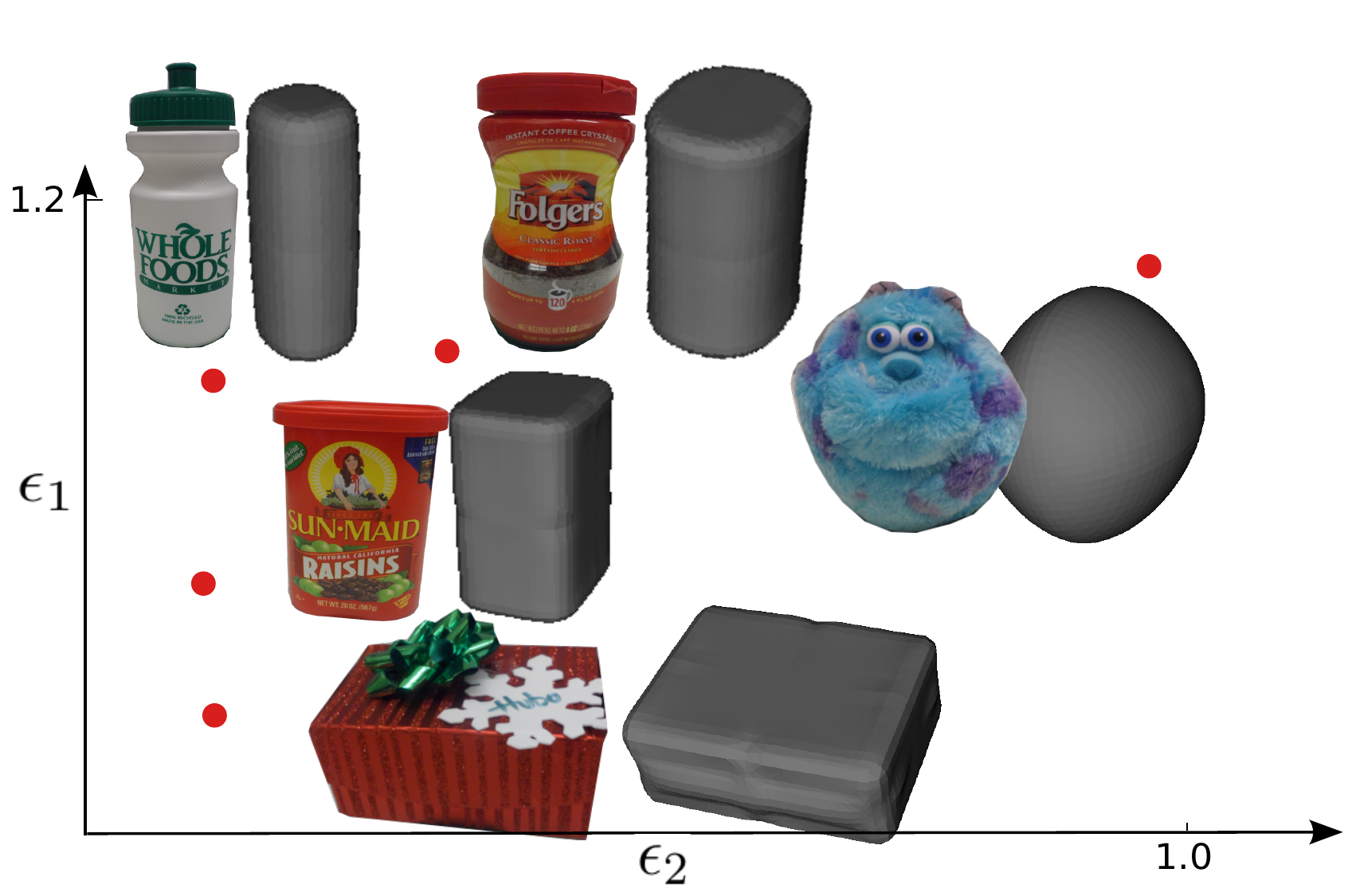} 
\caption{Examples of superquadrics with different shapes. A variety of
shapes can be represented using a small number of parameters. }
\label{fig:exampleSQ}
\end{center}
\end{figure}

\fref{fig:exampleSQ} shows different geometric shapes corresponding
to superquadrics with different values for $\epsilon_{1}$ and $\epsilon_{2}$. 

The superquadric approach turns the pointcloud-based representation into
a parametric representation, which can be much more efficiently used during 
grasp synthesis. Calculations of principal-axes, normals and other features
are much faster and less susceptible to noise.

\subsection{Generating Valid Candidates}
\label{subsec:validSolutions}

Once the shape of \Object~is approximated, we can proceed to generate candidate grasps \grasp{i}
using a simulation of the robot, and the object \Object, whose mesh is reconstructed
by using the superquadric parameters found in the previous section.

The candidates grasps must be kinematically feasible to execute with \Object~ located at
both start and goal conditions (\Tf{w}{s} and \Tf{w}{g}). Our approach accomplish this
with  \aref{alg:generateCandidates}. First, we set \Object~at its goal pose \Tf{w}{g}
and generate a set of kinematically feasible grasps for it (\GraspSet{}). 
Next, we set \Object~at its start
pose \Tf{w}{s}. Finally, we test each of the grasps 
\grasp{i} from \GraspSet{} in this scenario, discarding the grasps
for which there exist not a single IK solution. The surviving
grasps in \GraspSet{} are then grasps that can be executed for the object
\Object~at both \Tf{w}{s} and \Tf{w}{g}.

\begin{algorithm}
\DontPrintSemicolon
\KwIn{ \Hand, \Arm{}, \Tf{w}{s}, \Tf{w}{g}, \Object, \param }
\KwOut{Set of Candidate grasps \GraspSet{} }
\SetKwFunction{generateGrasps}{generate\_Grasps}
\SetKwFunction{setPose}{set\_Pose}
\SetKwFunction{erase}{.erase}
\SetKwFunction{existIKSol}{exist\_IK\_sol}
\SetKwFunction{FALSE}{false}
\BlankLine
\setPose{\Object,\Tf{w}{g}}\;
\tcc{Generate grasps with \Object ~at \Tf{w}{g}}
\GraspSet{} $\leftarrow$ \generateGrasps{\Hand,\Arm{},\Tf{w}{g},\Object, \param}\;
\setPose{\Object,\Tf{w}{s}}\;
\tcc{Discard grasps invalid with \Object ~at \Tf{w}{s}}
\ForEach{ \grasp{i} $\in$ \GraspSet{} }{
 \If{ \existIKSol{\grasp{i}, \Hand, \Arm} {\bf is} \FALSE  }{
   \GraspSet{}\erase{\grasp{i}}
 }
}\;
\Return \GraspSet{}\;
\caption{get\_Valid\_Candidates}
\label{alg:generateCandidates}
\end{algorithm}

\subsubsection{Grasp Generation at Goal Pose}
\label{subsub:graspGen}
The function \texttt{generate\_Grasps}, which we use
to produce grasps exploiting the shape parameters of \Object~is 
shown in \aref{alg:generateGrasps}. First, we uniformly sample
the surface of \Object. This is easily done by using the 
explicit equation defining the points in a superquadric and their 
corresponding normals:

\begin{equation}
\begin{bmatrix}
x \\ y \\ z
\end{bmatrix} =
\begin{bmatrix}
a\cos^{\epsilon_{1}}\eta \cos^{\epsilon_{2}}\omega \\
b\cos^{\epsilon_{1}}\eta \sin^{\epsilon_{2}}\omega \\
c\sin^{\epsilon_{1}}\eta 
\end{bmatrix} 
\quad \text{with} \quad
\begin{matrix}
\frac{\pi}{2} < \eta < \frac{\pi}{2} \\
\pi < \omega < \pi
\end{matrix}
\label{eq:expSQeq}
\end{equation} 

\begin{equation}
\begin{bmatrix}
n_{x} \\ n_{y} \\ n_{z}
\end{bmatrix} =
\begin{bmatrix}
\dfrac{1}{a}\cos^{2-\epsilon_{1}}\eta \cos^{2-\epsilon_{2}}\omega \\
\dfrac{1}{b}\cos^{2-\epsilon_{1}}\eta \sin^{2-\epsilon_{2}}\omega \\
\dfrac{1}{c}\sin^{2-\epsilon_{1}}\eta 
\end{bmatrix} 
\label{eq:normalSQeq}
\end{equation} 

Sampling uniformly $\omega$ and $\eta$ does not produce 
a uniform sampling of surface points due to the high nonlinearity
of the superquadrics equation. We use the method proposed by Pilu and
Fischer \cite{pilu1995equal} to obtain an evenly-spaced set of points 
and normals. 

To define a grasp we calculate the transformation of the hand \Hand{}
w.r.t. \Object (\Tf{o}{h}). We use the samples to generate this transformation
(lines 3 to 7 of \aref{alg:generateGrasps}). After positioning the TCP
 of the \Hand{}
at \Tf{w}{p}, we close the fingers. If there are not collisions with
the environment, we proceed to evaluate if there exists at least an
arm configuration that allows the hand to execute the grasp. If so, then
a corresponding grasp is stored.

\begin{algorithm}
\DontPrintSemicolon
\KwIn{ \Hand, \Arm{}, \Tf{w}{o}, \Object, \param}
\KwOut{A feasible set of grasps \GraspSet{} }
\SetKwFunction{sampleSQ}{sample\_SQ}
\SetKwFunction{smallestAxis}{smallest\_Axis}
\SetKwFunction{setHandTtcp}{setHand\_Tcp}
\SetKwFunction{closeHand}{close\_Hand}
\SetKwFunction{checkCollision}{check\_collision}
\SetKwFunction{existIKSol}{exist\_IK\_conf}
\SetKwFunction{Grasp}{Grasp}
\SetKwFunction{cross}{.cross}
\SetKwFunction{FALSE}{false}
\SetKwFunction{TRUE}{true}
\BlankLine
\tcc{\param = \{\ep{1},\ep{2},a,b,c\}  }
\SampleSet = \sampleSQ{\param} \;
\BlankLine
\ForEach{($p_{i}$,$n_{i}$) $\in$ $\SampleSet$}{
\tcc{p: TCP point in the hand \Hand}
  \Tf{o}{p}.trans = $p_{i}$\;
  \tcc{ z: Approach direction of \Hand}
  \Tf{o}{p}.z = -$n_{i}$\;
  \tcc{ x: Fingers closing direction }
  \Tf{o}{p}.x = \smallestAxis{a,b,c}\;
  \Tf{o}{p}.y = \Tf{o}{h}.z $\times$ \Tf{o}{h}.x\;
  \Tf{w}{p} = \Tf{w}{o}$\cdot$\Tf{o}{p}\;
  \tcc{h: Origin of hand \Hand}
  \Tf{w}{h} = \Tf{w}{p}$\cdot$\Tf{p}{h}\; 
  \setHandTtcp{\Hand, \Tf{w}{p}}\;
  \closeHand{\Hand}\;
  \If{ \checkCollision{\Hand} {\bf is} \FALSE }{
    \If{ \existIKSol{\Hand, \Arm{},\Tf{w}{h} } {\bf is} \TRUE }{
      \GraspSet{} $\leftarrow$ \Grasp{\Hand, \Tf{o}{p}$\cdot$\Tf{p}{h}}
    }
  }
}\;
\Return \GraspSet{}\;
\caption{GenerateGrasps(\Hand,\Arm,\Tf{w}{o},\Object,\param)}
\label{alg:generateGrasps}
\end{algorithm}

\begin{figure}[t]
\begin{center}
  \begin{tikzbox} {
  \begin{tikzpicture}[minimum width=0]
  	\tikzstyle{label} = [node distance = 1.7cm,text=blue]
    \tikzstyle{mycoord} = [node distance = 0cm]
    \tikzstyle{block} = [node distance = 3.5cm,rounded corners=.00cm,
    inner sep=.1cm, fill=white, minimum height=2em,minimum width=7em]
    \node[block,name=cover] {\includegraphics[width=0.4\textwidth]{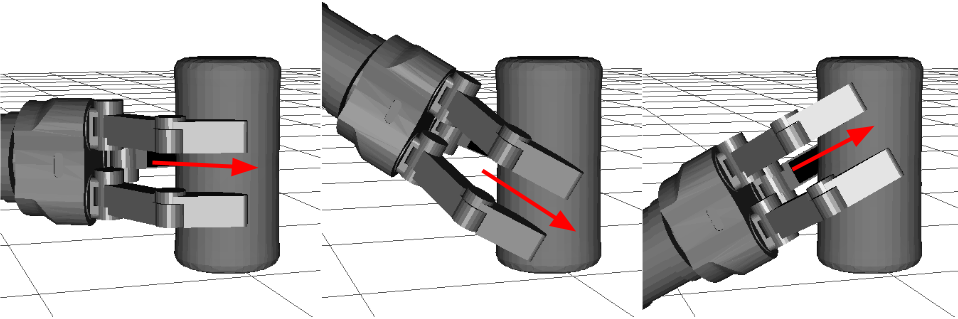} };
  \end{tikzpicture}
  }\end{tikzbox}
\caption{Generating grasps varying the approach direction by rotating the hand around
the local $x$ axis(pointing out the page): Left: Original. Middle,right: After rotating by $\pm\alpha$}
\label{fig:variatingGrasps}
\end{center}
\vspace{-2.0em}
\end{figure}

\aref{alg:generateGrasps} generates at most one grasp per each sampled point.
Optionally, we generated 2 additional possible grasps per each point by
rotating the hand an angle $\pm\alpha$ around the $x$ axis of \Tf{o}{p}. We 
added this since we noticed that, when executing the grasps on the physical robot,
a slight inclination usually made the grasp much easier to reach. In this paper we
used $\alpha=30^{o}$. An example of the variated grasps generated using $\alpha$
is shown in \fref{fig:variatingGrasps}.

\subsubsection{Validation at Start Pose}
\label{subsub:startPose}
Once a set of grasps feasible to execute on \Object~at \Tf{w}{g} is obtained, 
our algorithm discards the grasps that cannot be executed with \Object~at
\Tf{w}{s} (lines 4 to 6).

%% file: graspPrioritization.tex
Once a set of feasible grasps \GraspSet{} is generated, paths for reaching
and placing the object must be produced. A brute-force approach would be to exhaustively try each grasp in a random order until a solution is found. However, arm planning can be a time-consuming process, particularly when using sampling-based methods. It is therefore desirable to first evaluate grasps that are more likely to produce a solution. Since there is likely more than one solution
in \GraspSet{}, it is preferable to choose grasps such that the solution is quickly found. We propose to use \textit{situated grasp manipulability}
as a metric to prioritize the grasps and, hence, as a heuristic for guiding the search process. 

Manipulability($m$) measures how dexterous the end-effector
of a robotic arm is at a given joint configuration \conf{}. Initially
proposed by Yoshikawa \cite{yoshikawa1985manipulability}, 
\manip{\conf{}} is defined as:

\begin{equation}
m(\conf{}) = \sqrt{\begin{vmatrix}J(\conf{})J^{T}(\conf{})\end{vmatrix}}.
\end{equation}

Manipulability is typically defined for a single joint configuration.
In our scenario, we describe a situated grasp \grasp{i} for which multiple \conf{} might exist, due to redundancy. This naturally leads to 
the definition of \emph{situated grasp manipulability}($m_{g}$). Given a target object \Object~located at \Tf{w}{o}, and 
its corresponding grasp \grasp{i}, we define  $m_{g}$ as the 
average manipulability of a uniform set of collision-free 
arm configurations $\conf{i}$ that allow executing \grasp{i}:

\begin{equation}
m_{g} = \dfrac{1}{N} \sum_{i=1}^{N}m(\conf{i})
\label{eq:graspManip}
\end{equation}

\begin{figure}[h]
\begin{center}
  \begin{tikzbox} {
  \begin{tikzpicture}[minimum width=0]
  	\tikzstyle{label} = [node distance = 1.7cm,text=blue]
    \tikzstyle{mycoord} = [node distance = 0cm]
    \tikzstyle{block} = [node distance = 3.5cm,rounded corners=.00cm,
    inner sep=.1cm, fill=bgcolor, minimum height=2em,minimum width=7em]
    \node[block,name=cover] {\includegraphics[width=0.4\textwidth]{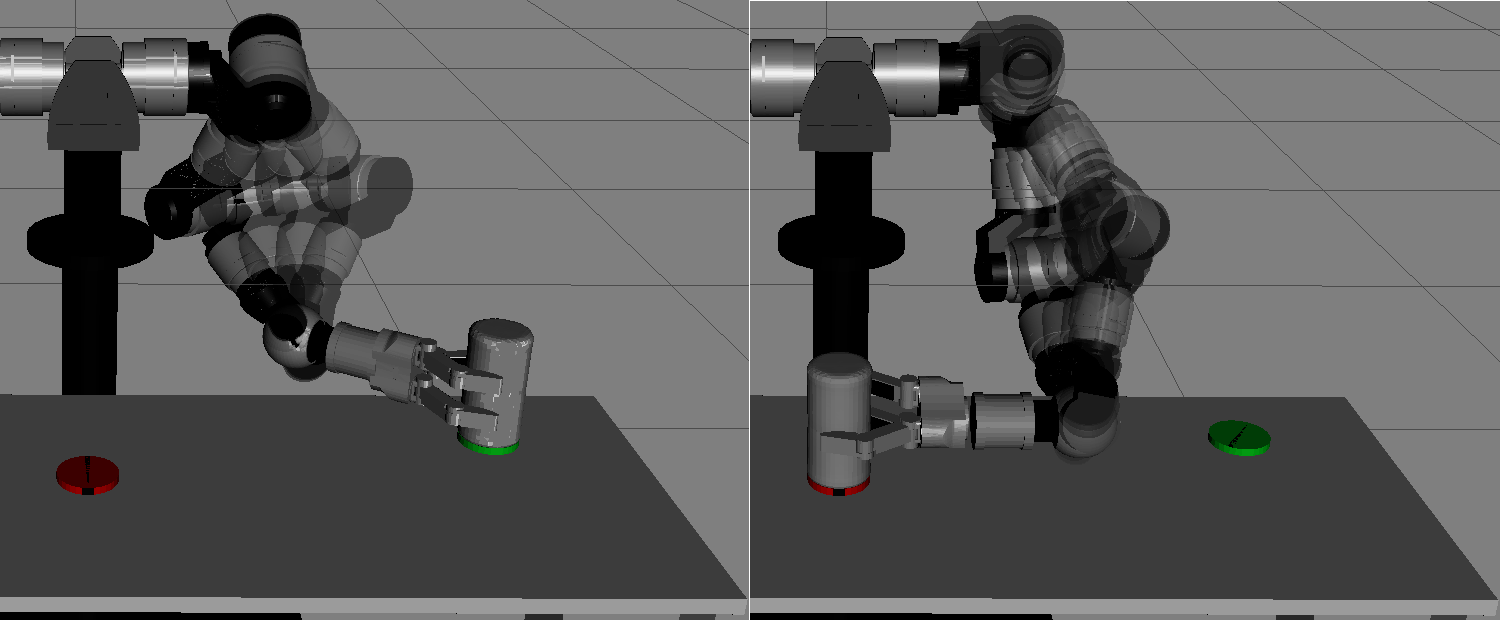} };
  \end{tikzpicture}
  }\end{tikzbox}
\caption{Examples of $m_{g}$ measured at \Tf{w}{s} and \Tf{w}{g}}
\label{fig:graspManip}
\end{center}
\vspace{-1.0em}
\end{figure}

Please note that $m_{g}$ depends on both \grasp{i}, \Tf{w}{o} and the
environment (for collisions) since only collision-free grasps that reach 
the object are considered. \fref{fig:graspManip} shows an example
of a pick-and-place task wherein the green and red markers indicate
the \Tf{w}{s} and \Tf{w}{g}. In this case, $m_{g}$ at \Tf{w}{g}
is bigger than \Tf{w}{s} (where $N_{s}=76$ and $N_{g}=108$ are the number
of IK solutions for both situations). When the object is at \Tf{w}{g}, the arm movement requires less effort.

From the shown example, it becomes evident that for a pick-and-place manipulation problem, there are at least two possible metrics to use per grasp \grasp{i}: $m_{g}$ measured either at \Tf{w}{s} or \Tf{w}{g}. Choosing the first option means that we prioritize grasps in which the pick phase is executed comfortably (\Tf{w}{s}), whereas by choosing the latter, we favor grasps in which the arm configuration used at placing the object (\Tf{w}{g}) is more relaxed. In section \ref{sec:experiments}, we present the results of experiments comparing these two metrics and an additional control measure to analyze their performance and choose which one is best suited for our problem.


%% file: comparativeStudies.tex
In this section, we perform a set of experiments in simulation and on the real robot in order to evaluate the introduced manipulation planning algorithm. The simulation experiments are used to analyze the situated grasp manipulability using a large number of trials. Experiments on the real robots are performed to show the generation of manipulations based on task goals and previously unknown objects. Generation of superquadric object models was performed on the spot within ~1 second. 

\subsection{Simulation Experiments}
In this experiment, we consider three alternatives:

\begin{itemize}
\item{Measure $m_{g}$ for grasp situated at \Tf{w}{s} }
\item{Measure $m_{g}$ for grasp situated at \Tf{w}{g} }
\item{Average of both measures above}
\end{itemize}

We use 3 measures to compare the performance 
of the 3 evaluated metrics.
\begin{itemize}
\item{\textit{First success:} The main goal of the metrics evaluated is
to prioritize the grasps such that the first one tried is the most likely to succeed.
This metric indicates the number of times that a solution is found by evaluating
only the grasp with the biggest value for the evaluated metric.}
\item{\textit{Planning time:} It measures the average planning time of the
successing grasps. The planning time is the total time to plan a reach and transport
path for the given grasp.}
\item{\textit{Path length:} It indicates the number of steps required for
the pick-and-place solution. The step length is a normalized value in joint space, so
this metric compare the paths in configuration space.}
\end{itemize}

\begin{figure}[t]
\begin{center}
  \begin{tikzpicture}[minimum width=0]
  	\tikzstyle{label} = [node distance = 1.7cm,text=blue]
    \tikzstyle{mycoord} = [node distance = 0cm]
    \tikzstyle{block} = [node distance = 3.5cm,rounded corners=.00cm,
    inner sep=.1cm, fill=bgcolor, minimum height=2em,minimum width=7em]
    \node[block,name=cover] {\includegraphics[width=0.4\textwidth]{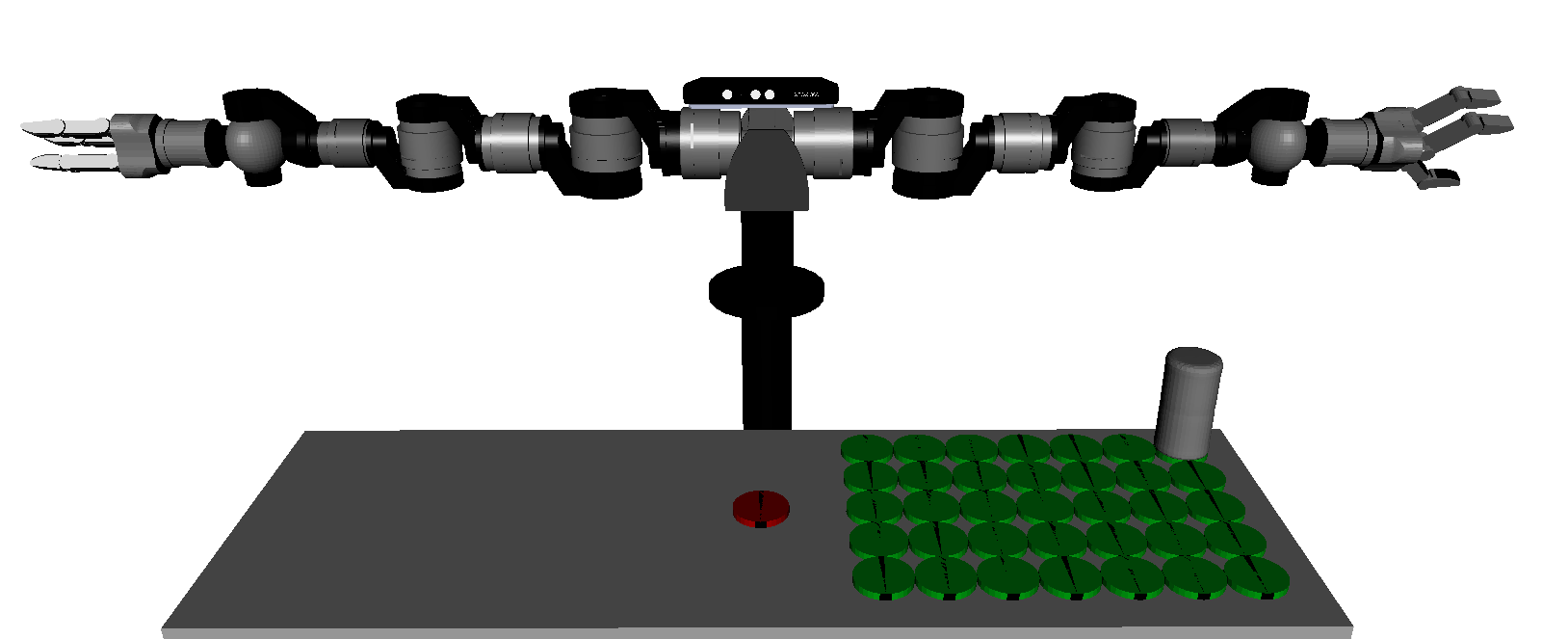} };
  \end{tikzpicture}
\caption{Setup for unimanual evaluation experiments}
\label{fig:unimanualSetup}
\end{center}
\vspace{-1.0em}
\end{figure}

The scenario we used is depicted in \fref{fig:unimanualSetup}. We fix the \Tf{w}{g} 
to the middle of the table (red marker) and vary the start pose \Tf{w}{s} to 35 
positions, each separated 0.1 m (green markers). We devised 2 kind of experiments: In
the first, \Tf{w}{s} and \Tf{w}{g} have the same orientation, with only the position
being changed (35 scenarios). In the second case, \Tf{w}{s} presents a rotation around the Z axis
w.r.t. \Tf{w}{g} in the interval $[0,2\pi]$ at each $\dfrac{\pi}{4}$ steps, so
in total $35\times8=280$ scenarios are tested.

We used an standard IK-BiRRT to perform the arm planning. To account for its randomness,
the results presented  in \tref{tab:evalNoRotation} and \tref{tab:evalRotation} are an average of 5 runs per each experiment. 

\begin{table}[h]
\centering
\caption{Evaluation with no rotation change}
\label{tab:evalNoRotation}
\begin{tabular}{l c c c}
\topline
\headcol \textbf{Metric Type} & \textbf{Path Steps} & \textbf{Planning Time} & \textbf{Success} \\
\midline
$m_{g}$ at \Tf{w}{s} & 82.92 & 2.17 & 21.8/35 \\
\cmidrule{1-4}
$m_{g}$ at \Tf{w}{g} & 89.28 & 2.218 & {\color{blue}33/35} \\
\cmidrule{1-4}
Avg. $m_{g}$ & 92.92 & 2.29 & 31.4/35 \\
\bottomline
\end{tabular}
\vspace{-0.0em}
\end{table}

\begin{table}[h]
\centering
\caption{Evaluation with rotation change}
\label{tab:evalRotation}
\begin{tabular}{l c c c}
\topline
\headcol \textbf{Metric Type} & \textbf{Path Steps} & \textbf{Planning Time} & \textbf{Success} \\
\midline
$m_{g}$ at \Tf{w}{s} & 100.7 & 4.70 & 255/278 \\
\cmidrule{1-4}
$m_{g}$ at \Tf{w}{g} & 99.6 & 4.49 & {\color{blue}275/278} \\
\cmidrule{1-4}
Avg. $m_{g}$ & 100.4 & 3.83 & 260/278 \\
\bottomline
\end{tabular}
\vspace{-0.0em}
\end{table}

From the tables, we can observe that using $m_{g}$ evaluated at \Tf{w}{g} produces
the best results in terms of success at the first trial, whereas \Tf{w}{s} present
the worst results. The average path length for the general case of \tref{tab:evalRotation}
is rather similar for the 3 cases. Regarding planning times, the average $m_{g}$ gives
better results.

Given the results presented, we chose to use the $m_{g}$ at \Tf{w}{g}. Its next best 
competitor (the avg. $m_{g}$) was not considered since in order to calculate it, the $m_{g}$
at both \Tf{w}{s} and \Tf{w}{g} must be calculated, which increases the computation time
(for the examples presented, the computation time of $m_{g}$ was ~2 seconds). Given
that the advantage of planning time is not significant, we chose $m_{g}$ at \Tf{w}{g}.



\subsection{Robot Experiments}
\begin{figure}[h]
\begin{center}
\includegraphics[width=0.45\textwidth]{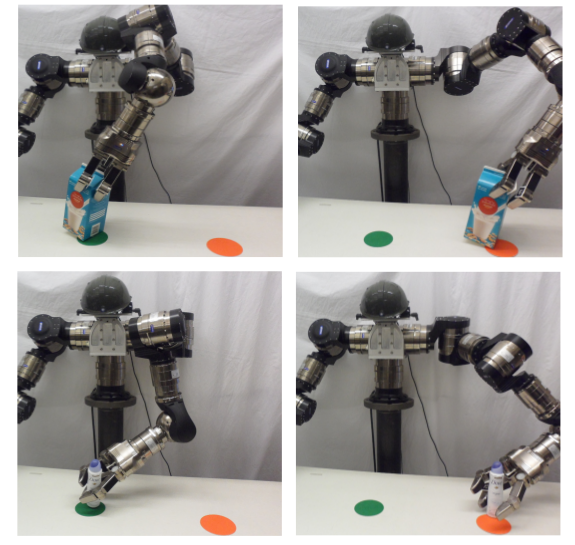} 
\caption{Two examples of a pick-and-place without constraints. The 
robot can identify suitable grasp for novel objects using the superquadric
representation.}
\label{fig:noconstraint}
\end{center}
\vspace{-1.0em}
\end{figure}

Next, we perform a set of experiments on the real robot. All performed experiments are
pick-and-place tasks. However, in some tasks we add environmental constraints at the
goal location which limit the range of applicable grasps. \fref{fig:noconstraint} depcits
two trials without any environmental constraint. The robot has to pick an object at the
starting location (green) and move it successfully to the goal location (orange). The robot has no prior knowledge of the object and needs to extract shape information from a single depth image produced by a depth sensor mounted in the head. As can be seen in the figure,
the grasp direction and the hand shape is adapted to suit the object.

\begin{figure*}[!t]
\begin{center}
\begin{tabular}{ccc}
\subfloat{\includegraphics[width=0.3\textwidth]{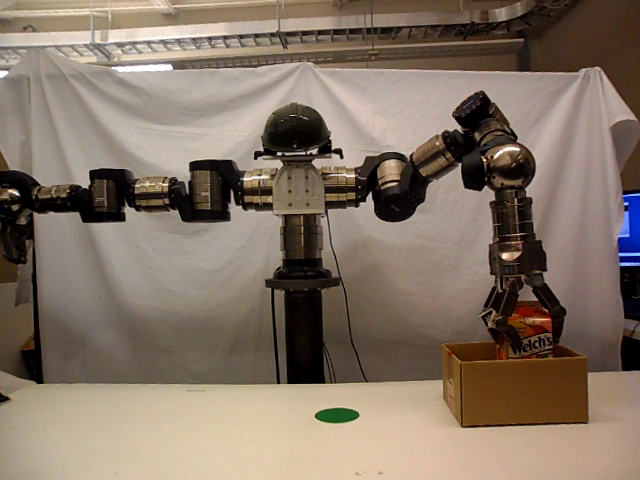}} \hspace*{-0.9em} &
\subfloat{\includegraphics[width=0.3\textwidth]{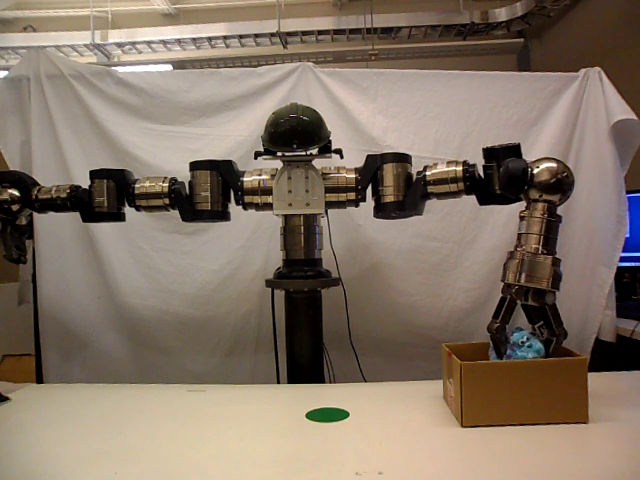}} \hspace*{-0.9em} &
\subfloat{\includegraphics[width=0.3\textwidth]{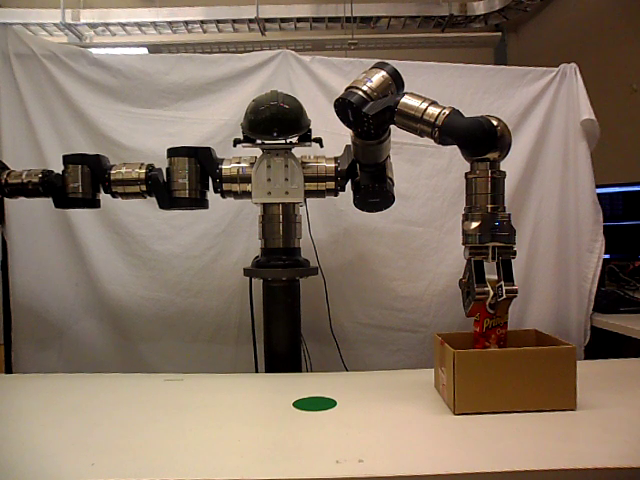}} \hspace*{-0.9em}\\ 
\subfloat{\includegraphics[width=0.3\textwidth]{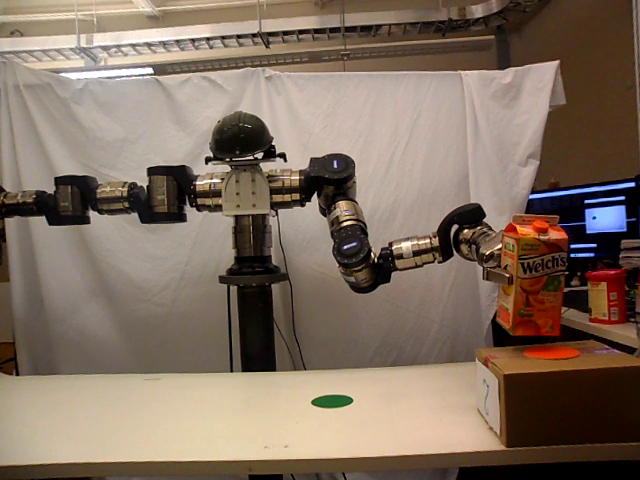}} \hspace*{-0.9em} &
\subfloat{\includegraphics[width=0.3\textwidth]{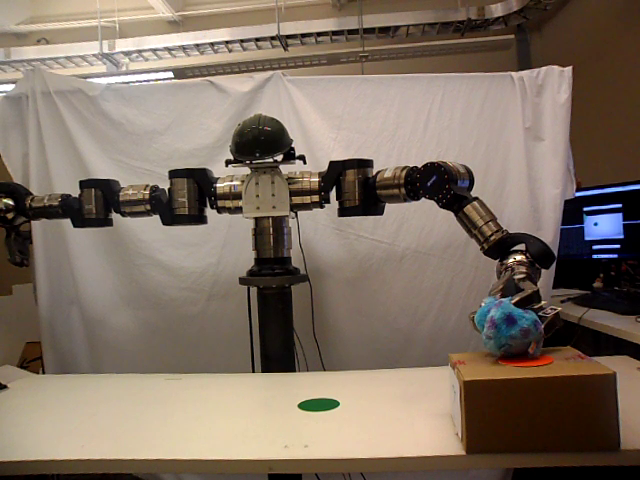}} \hspace*{-0.9em} &
\subfloat{\includegraphics[width=0.3\textwidth]{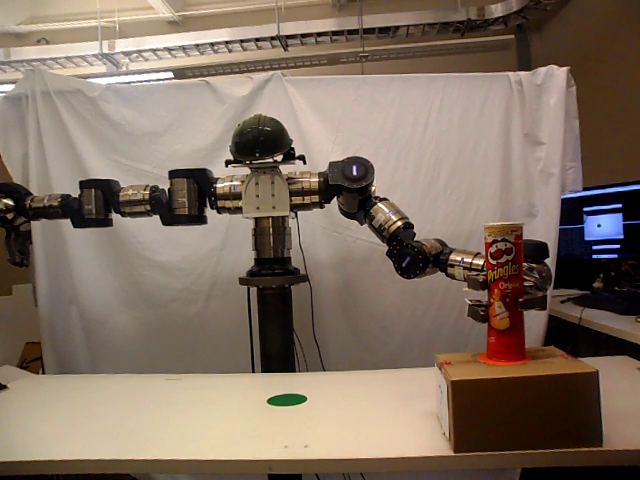}} \hspace*{-0.9em}\\ 
\subfloat{\includegraphics[width=0.3\textwidth]{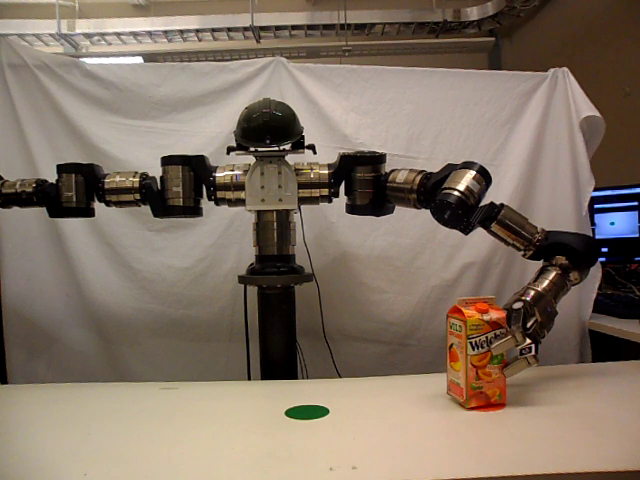}} \hspace*{-0.9em} &
\subfloat{\includegraphics[width=0.3\textwidth]{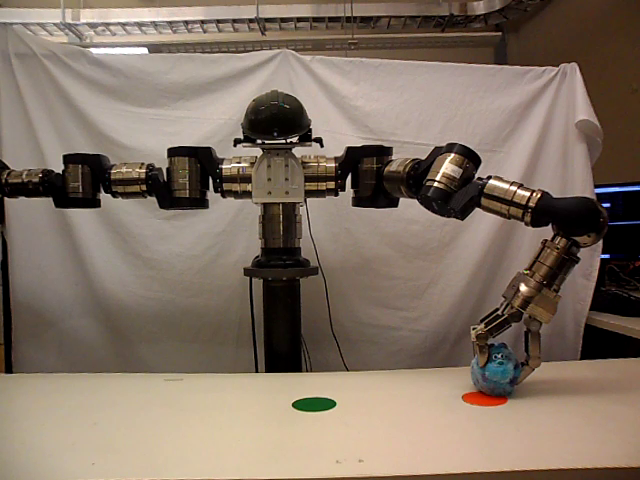}} \hspace*{-0.9em} &
\subfloat{\includegraphics[width=0.3\textwidth]{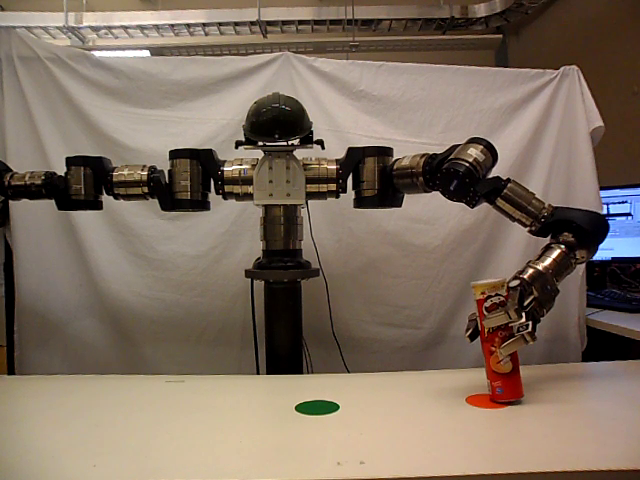}} \hspace*{-0.9em}
\end{tabular}
\end{center}
\caption{Final grasp configuration during manipulation with goal constraints (top and middle) and without goal constraints (bottom).}
\label{fig:withconstraint}
\end{figure*}

A different set of experiments can be seen in \fref{fig:withconstraint}. Here, environmental constraints at the goal are introduced. In the top row, the object has to be placed in a box. Accordingly, the robot has to choose a grasp that allows it to place the
object in the box without colliding with it. Hence, the selected grasps are mostly from above. The middle row show a different scenario, in which the object has to be placed on a box which is farther away. Choosing the wrong approach direction, e.g. from above, would prevent the robot from successfully finishing the manipulation process, due to workspace limitations. The bottom row shows normal runs without any environmental constraints.

%% file: conclusion.tex
In this paper, we introduced a new method for manipulation planning with task constraints. Given a previously unknown object and goals of the task, the method synthesizes online a grasp that facilitates task completion. Planning and grasp synthesis are effectively merged to efficiently produce manipulation sequences. Object acquisition, representation, grasp synthesis and planning can be performed within a couple of seconds, i.e., ~2-5 seconds, for the presented examples. We showed how superquadrics and a new heuristic, i.e., the situated grasp manipulability can be used towards this end. These properties, hence, allow a robot to perform successful, goal-driven manipulation tasks in new environments without resorting to prior 3D models of the object or pre-calculated grasps.

While superquadrics can be efficiently calculated, they lack accuracy when representing complex shapes and objects. In this paper, we showed that many objects, in particular household objects can be represented by superquadrics. In our future work, we want to explore extensions of this representation that can model a larger set of objects. In particular, we want to build upon our previous research on the identification of rotational and linear extrusions \cite{huaman2015sq} to represent more complex shapes. In addition, we want to verify the introduced planning approach on longer manipulation sequences.